\documentclass[10pt,twocolumn,letterpaper]{article}

\usepackage{cvpr}
\usepackage{times}
\usepackage{epsfig}
\usepackage{graphicx}
\usepackage{amsmath}
\usepackage{amssymb}
\usepackage{xcolor}

\usepackage{booktabs}
\usepackage{tabulary}
\usepackage{array, multirow}
\usepackage{rotating}
\usepackage{cite}
\usepackage{algorithm}
\usepackage[noend]{algpseudocode}

\usepackage[pagebackref=true,breaklinks=true,letterpaper=true,colorlinks,bookmarks=false]{hyperref}

\cvprfinalcopy 

\begin{document}

\title{Understanding the Limitations of CNN-based Absolute Camera Pose Regression}

\author{Torsten Sattler$^1$
\and
Qunjie Zhou$^2$
\and
Marc Pollefeys$^{3,4}$
\and
Laura Leal-Taix\'{e}$^2$
\and
$^1$Chalmers University of Technology \quad $^2$TU Munich \quad $^3$ETH Z\"{u}rich \quad  $^4$Microsoft
}

\maketitle

\newcommand{\Eq}{Eq.\xspace}
\newcommand{\Eqs}{Eqs.\xspace}
\newcommand{\Fig}{Fig.\xspace}
\newcommand{\Figs}{Figs.\xspace}
\newcommand{\Sec}{Sec.\xspace}
\newcommand{\Tab}{Tab.\xspace}
\newcommand{\Tabs}{Tabs.\xspace}

\newcommand{\NEW}[1]{\textcolor{red}{#1}}
\newcommand{\NEWC}[1]{\textcolor{blue}{#1}}
\newcommand{\PAR}[1]{\vskip4pt \noindent{\bf #1~}}

\newcommand{\lau}[1]{\textcolor{magenta}{#1}\xspace}		
\newcommand{\ts}[1]{\textcolor{blue}{#1}\xspace}                     
\newcommand{\qj}[1]{\textcolor{olive}{Qj: #1}\xspace}                     
\newcommand{\mapo}[1]{\textcolor{red}{Marc: #1}\xspace}                     

\definecolor{darkgreen}{RGB}{31, 150, 36}
\definecolor{darkred}{RGB}{149, 8, 20}
\definecolor{brightcyan}{RGB}{45, 255, 254}
\definecolor{deepmagenta}{rgb}{0.8, 0.0, 0.8}

\begin{abstract}
Visual localization is the task of accurate camera pose estimation in a known scene. It is a key problem in computer vision and robotics, with applications including self-driving cars, Structure-from-Motion, SLAM, and Mixed Reality. Traditionally, the localization problem has been tackled using 3D geometry. Recently, end-to-end approaches based on convolutional neural networks have become popular. These methods learn to directly regress the camera pose from an input image. However, they do not achieve the same level of pose accuracy as 3D structure-based methods. To understand this behavior, we develop a theoretical model for camera pose regression. We use our model to predict failure cases for pose regression techniques and verify our predictions through experiments. We furthermore use our model to show that pose regression is more closely related to pose approximation via image retrieval than to accurate pose estimation via 3D structure. A key result is that current approaches do not consistently outperform a handcrafted image retrieval baseline. This clearly shows that additional research is needed before pose regression algorithms are ready to compete with structure-based methods.
\vspace{-6pt}
\end{abstract}

\vspace{-12pt}
\section{Introduction}
\vspace{-3pt}
Visual localization algorithms enable a camera to determine its absolute pose, \ie, its position and orientation, in a scene. 
Localization thus is a core component for intelligent systems such as self-driving cars~\cite{Haene2017IMAVIS} or other robots~\cite{Lim2015IJRR}, and for Augmented and Mixed Reality applications~\cite{Middelberg2014ECCV,Castle08ISWC}. 

State-of-the-art algorithms for localization follow a 3D structure-based approach~\cite{Sattler2017PAMI,Svarm2017PAMI,Brachmann2017CVPR,Brachmann2018CVPR,Taira2018CVPR,Schoenberger2018CVPR,Cavallari2017CVPR,Meng2017IROS}. 
They first establish correspondences between pixels in a test image and 3D points in the scene. 
These 2D-3D matches are then used to estimate the camera pose by applying an $n$-point-pose (PnP) solver~\cite{Kneip2011CVPR,Kukelova2013ICCV,Larsson2017ICCV,Albl2015CVPR,Larsson2018CVPR} inside a RANSAC~\cite{Fischler81CACM,Lebeda2012BMVC,Chum08PAMI,Raguram2013PAMI} loop. 
Traditionally, the first stage is based on matching descriptors extracted in the test image against descriptors associated with the 3D points. 
Alternatively, machine learning techniques can be used to directly regress 3D point positions from image patches~\cite{Brachmann2017CVPR,Brachmann2018CVPR,Cavallari2017CVPR,Massiceti17CVPR,Meng2018IROS,Meng2017IROS,Shotton2013CVPR}.

In recent years, absolute pose regression (APR) approaches to visual localization have become popular~\cite{Kendall2015ICCV,Kendall2016ICRA,Kendall2017CVPR,Walch2017ICCV,Cai2018BMVC,Melekhov2017ICCVW,Naseer2017IROS,Radwan2018RAL,Valada2018ICRA,Wu2017ICRA,Brahmbhatt2018CVPR}. 
Rather than using machine learning only for parts of the localization pipeline, \eg, local features~\cite{Yi2016ECCV,Schoenberger2018CVPR}, outlier filtering~\cite{Toft2018ECCV,Yi2018CVPR}, or scene coordinate regression~\cite{Brachmann2018CVPR,Meng2017IROS}, these approaches aim to learn the full localization pipeline. 
Given a set of training images and their corresponding poses, APR techniques train Convolutional Neural Networks (CNNs) to directly regress the camera pose from an image. 
APR techniques are computationally efficient, given a powerful enough GPU, as only a single forward pass through a CNN is required. 
Yet, they are also significantly less accurate than structure-based methods~\cite{Walch2017ICCV,Schoenberger2018CVPR,Brachmann2018CVPR}. 
In addition, updating the map, \eg, when adding new data, requires expensive retraining of the CNN. %

Rather than proposing a new APR variant to try to close the pose accuracy gap to structure-based methods, this paper focuses on understanding APR techniques and their performance. 
To this end, we make the following contributions: 
\textbf{i}) We develop a theoretical model for absolute pose regression (Sec.~\ref{sec:theory}). 
To the best of our knowledge, ours is the first work that aims at looking at the inner workings of APR techniques. 
Based on this model, we show that APR approaches are more closely related to approximate pose estimation via image retrieval (Sec.~\ref{sec:practical_experiments}) than to accurate pose estimation via 3D geometry (Sec.~\ref{sec:generalization}). 
\textbf{ii}) Using our theory, we show both theoretically and through experiments that there is no guarantee that APR methods, unlike structure-based approaches, generalize beyond their training data (Sec.~\ref{sec:generalization}). 
\textbf{iii}) Given the close relation between APR and image retrieval, we show that current APR approaches are much closer in performance to a handcrafted retrieval baseline~\cite{Torii15CVPR} than to structure-based methods. 
We show that no published single image pose regression approach is able to consistently outperform this baseline. 
This paper thus introduces a highly necessary sanity check for judging the performance of pose regression techniques. 

In summary, this work closes an important gap in the understanding of absolute pose regression methods to visual localization: 
It clearly demonstrates their short-comings and more clearly positions them against other ways to approach the visual localization problem. 
Overall, we show that a significant amount of research is still necessary before absolute pose regression techniques can be applied in practical applications that require accurate pose estimates.

\begin{figure*}[t!]
\begin{center}
\includegraphics[width=0.85\linewidth]{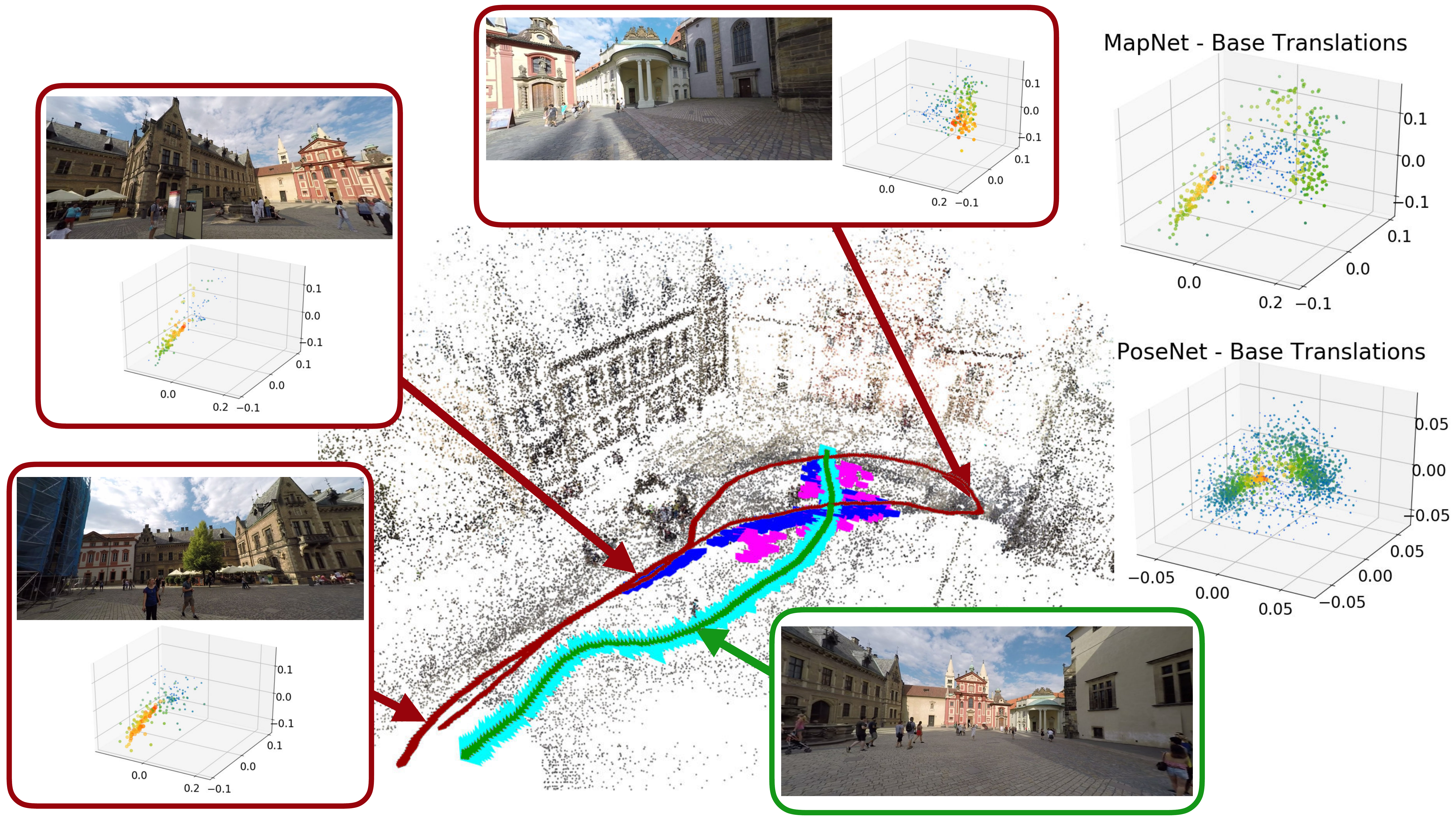}%
\end{center}%
\vspace{-6pt}
\caption{Visualization of the base translations $\{\mathbf{c}_j\}$ learned by PoseNet~\cite{Kendall2015ICCV,Kendall2017CVPR} and MapNet~\cite{Brahmbhatt2018CVPR}. Each point corresponds to one base translation. The scale of the base translations is in meters. We show the combinations of base translations for some training images for MapNet. The weight a translation received in Eq.~\ref{eq:linear_pose_combination} for a single image, respectively all images (on the right of the figure), is indicated by colors and point sizes, with warm colors and large points for translations with a large coefficient.  The training and test trajectory are shown in \textcolor{darkred}{red} and \textcolor{darkgreen}{green}. The test predictions by PoseNet and MapNet and Active Search~\cite{Sattler2017PAMI} are shown in \textcolor{blue}{blue}, \textcolor{deepmagenta}{purple}, and  
\textcolor{cyan}{cyan}, respectively.}%
\label{fig:example_bases}
\end{figure*}

\vspace{-6pt}
\section{Related Work}
\label{sec:related_work}
\vspace{-3pt}
\PAR{Structure-based localization} approaches rely on 2D-3D matches between 2D pixel positions and 3D scene coordinates for pose estimation. 
These matches are established by descriptor matching~\cite{Donoser14CVPR,Sattler2017PAMI,Svarm2017PAMI,Zeisl2015ICCV,Taira2018CVPR,Schoenberger2018CVPR,Li2012ECCV,Liu2017ICCV} or by regressing 3D coordinates from pixel patches~\cite{Brachmann2017CVPR,Brachmann2018CVPR,Cavallari2017CVPR,Meng2017IROS,Brachmann2016CVPR,Massiceti17CVPR,Meng2018IROS,Shotton2013CVPR,Guzman14CVPR}. 
Descriptor-based methods handle city-scale scenes~\cite{Liu2017ICCV,Zeisl2015ICCV,Svarm2017PAMI,Li2012ECCV} and run in real-time on mobile devices~\cite{Arth09ISMAR,Lim2015IJRR,Lynen2015RSS,Middelberg2014ECCV}. 
3D coordinate regression methods currently achieve a higher pose accuracy  at small scale, but have not yet been shown to scale to larger scenes~\cite{Taira2018CVPR,Brachmann2018CVPR}. 

\PAR{Image retrieval} is typically used for place recognition\cite{Chen2011CVPR,Arandjelovic2014ACCV,Arandjelovic16CVPR,Sattler2016CVPR,Torii2011ICCVW,Torii15CVPR,Zamir10ECCV,Weyand2016ECCV}, \ie, for determining which part of a scene is visible in a given image. 
State-of-the-art approaches use compact image-level descriptors to enable efficient and scalable retrieval~\cite{Arandjelovic16CVPR,Radenovi2018TPAMI,Torii15CVPR}. 
Image retrieval can be used for visual localization by approximating the pose of a test image by the pose of the most similar retrieved image. 
More precise estimates can be obtained by using feature matches between the test image and the retrieved images for relative pose estimation~\cite{Camposeco2018CVPR,Zhang06TDPVT,Zheng2015ICCV}. 
Image retrieval has also been used as part of structure-based approaches~\cite{Cao13CVPR,Irschara09CVPR,Sattler2015ICCV}.

\PAR{Absolute camera pose regression (APR)} approaches train CNNs to regress the camera pose of an input image~\cite{Kendall2015ICCV,Kendall2016ICRA,Kendall2017CVPR,Melekhov2017ICCVW,Walch2017ICCV,Wu2017ICRA,Naseer2017IROS,Brahmbhatt2018CVPR,Radwan2018RAL,Valada2018ICRA}, thus representing the scene implicitly by the weights of the networks. 
They all follow the same pipeline: Features are extracted using a base network, \eg, VGG~\cite{Simonyan15ICLR} or ResNet~\cite{Kaiming2016CVPR}, which are then embedded into a high-dimensional space. 
This embedding is then used to regress the camera pose in the scene. 
Existing approaches mainly differ in the underlying base architecture and the loss function used for training, \eg, using a weighted combination of position and orientation errors~\cite{Kendall2015ICCV,Brahmbhatt2018CVPR,Walch2017ICCV}, geometric reprojection errors~\cite{Kendall2017CVPR}, or adding visual odometry constraints~\cite{Brahmbhatt2018CVPR,Radwan2018RAL,Valada2018ICRA}. 
\cite{Wu2017ICRA,Naseer2017IROS} extend the set of training images with synthetic data. 
\cite{Cai2018BMVC,Kendall2016ICRA} also reason about the uncertainty of the estimated poses. 
Rather than using a single image, \cite{Clark2017CVPR,Radwan2018RAL,Brahmbhatt2018CVPR,Valada2018ICRA} propose methods based on localizing sequences of images. 

Recent results show that APR methods are significantly less accurate than structure-based methods~\cite{Walch2017ICCV,Brachmann2018CVPR,Meng2017IROS}. 
This paper aims to understand these results by developing a theoretical model for APR. 
Based on this model, we show that, in contrast to structure-based methods, APR approaches struggle to generalize beyond their training data or might not generalize at all.
Furthermore, we show that APR techniques are inherently closer related to image retrieval than to structure-based methods and that current APR algorithms do not consistently outperform a retrieval baseline. 

\PAR{Relative camera pose regression (RPR)} approaches predict the pose of a test image relative to one or more training images rather than in absolute scene coordinates~\cite{Balntas2018ECCV,Laskar2017ICCVW,Melekhov2017ICAC,Saha2018BMVC}. 
The prediction is again handled by a CNN trained for regression. 
Relevant training images can be found using an explicit image retrieval step~\cite{Balntas2018ECCV,Laskar2017ICCVW} or by implicitly representing the images in the CNN~\cite{Saha2018BMVC}. 
APR is an instance-level problem, \ie, APR techniques need to be trained for a specific scene. 
In contrast, RPR is a more general problem and RPR methods can be trained on multiple scenes~\cite{Balntas2018ECCV,Laskar2017ICCVW}.

In this paper, we use our theory of APR to show that there is an inherent connection to RPR. 
We also show that, while being are among the best-performing end-to-end localization approaches, current RPR techniques also do not consistently outperform an image retrieval baseline.

\section{A Theory of Absolute Pose Regression}
\label{sec:theory}
The purpose of this section is to develop a theoretical model for absolute camera pose estimation methods such as PoseNet~\cite{Kendall2015ICCV,Kendall2016ICRA,Kendall2017CVPR} and its variants~\cite{Walch2017ICCV,Naseer2017IROS,Brahmbhatt2018CVPR,Wu2017ICRA}. 
Our theory is not tied to a specific network architecture but covers the family of architectures used for pose regression. 
Based on this theory, Sec.~\ref{sec:generalization} compares absolute pose regression and structure-based methods, using experiments to support our model. 
Sec.~\ref{sec:practical_experiments} then uses the theory to show the inherent similarities between pose regression and image retrieval.

\PAR{Notation}
Let $\mathcal{I}$ be an image taken from a camera pose $\mathbf{p}_\mathcal{I} = (\mathbf{c}_\mathcal{I}, \mathbf{r}_\mathcal{I})$.
Here, $\mathbf{c}_\mathcal{I} \in \mathbb{R}^3$ is the camera position and $\mathbf{r}_\mathcal{I}$ is the camera orientation.  
There are multiple ways to represent the orientation, \eg, as a 4D unit quaternion~\cite{Kendall2015ICCV,Walch2017ICCV} or its logarithm~\cite{Brahmbhatt2018CVPR}, or as a 3D vector representing an angle and an axis~\cite{Ummenhofer2017CVPR,Balntas2018ECCV}. 
The exact choice of representation is not important for our following analysis. 
Without loss of generality, we thus simply represent the orientation as a $r$-dimensional vector $\mathbf{r}_\mathcal{I} \in \mathbb{R}^r$. 
Absolute camera poses are thus represented as points in $\mathbb{R}^{3 + r}$.

\PAR{Absolute pose regression.}
Given a test image $\mathcal{I}$, the task of absolute camera pose regression is to predict the pose from which the image was taken. 
This pose is defined with respect to a given scene coordinate frame. 
To solve this tasks, algorithms for absolute camera pose regression learn a visual localization function $L(\mathcal{I}) = \hat{\mathbf{p}}_\mathcal{I}$, where $\hat{\mathbf{p}}_\mathcal{I} = (\hat{\mathbf{c}}_\mathcal{I}, \hat{\mathbf{r}}_\mathcal{I})$ is the camera pose predicted for image $\mathcal{I}$.
In the following, we will focus on methods that represent the function $L$ via a convolutional neural network (CNN)~\cite{Kendall2015ICCV,Kendall2016ICRA,Kendall2017CVPR,Walch2017ICCV,Brahmbhatt2018CVPR}. 

Absolute camera pose regression is an instance level problem. 
Thus, CNN-based methods for absolute pose regression use a set of images of the scene, labeled with their associated camera poses, as training data. 
Additional image sequences without pose labels might also be used to provide additional constraints~\cite{Brahmbhatt2018CVPR,Valada2018ICRA}. 
The training objective is to minimize a loss $\mathcal{L}(\hat{\mathbf{p}}_\mathcal{I}, \mathbf{p}_\mathcal{I})$ enforcing that the predicted pose $\hat{\mathbf{p}}_\mathcal{I}$ is similar to the ground truth pose $\mathbf{p}_\mathcal{I}$. 
The precise formulation of the loss is not important for our analysis.

\PAR{A theory of absolute pose regression.} 
We divide absolute pose regression via a CNN into three stages: 
The first stage, representing a function $F(\mathcal{I})$, extracts a set of features from the image. 
This stage is typically implemented using the fully convolutional part of a CNN such as  VGG~\cite{Simonyan15ICLR} or ResNet~\cite{Kaiming2016CVPR}. %
The second stage computes a (non-linear) embedding $E(F(\mathcal{I}))$ of the features into a vector $\mathbf{\alpha}^\mathcal{I}=(\alpha^\mathcal{I}_1, \dots, \alpha^\mathcal{I}_n)^T \in \mathbb{R}^n$ in a high-dimensional space. %
This embedding typically corresponds to the output of the second-to-last layer in a pose regression method. 
The last stage performs a linear projection from the embedding space into the space of camera poses. 
This third stage corresponds to the last (fully-connected) layer in the network. 
This three stage model covers all PoseNet-like approaches that have been published so far. 

Treating the first two stages as a single network, we can write the trained visual localization function $L$ as 
\begin{eqnarray}
L(\mathcal{I}) & = & \mathbf{b} + \mathtt{P} \cdot E(F(\mathcal{I})) \nonumber \\ 
& = & \mathbf{b} + \mathtt{P} \cdot \left(\alpha^\mathcal{I}_1, \dots, \alpha^\mathcal{I}_n\right)^T \enspace ,
\end{eqnarray}
where $\mathtt{P} \in \mathbb{R}^{(3+r) \times n}$ is a projection matrix and $\mathbf{b} \in \mathbb{R}^{3+r}$ is a bias term. 
The output of $L(\mathcal{I})$ is an estimate $\hat{\mathbf{p}}_\mathcal{I} = (\hat{\mathbf{c}}_\mathcal{I}, \hat{\mathbf{r}}_\mathcal{I})$ of the image's camera pose. %
Let $\mathbf{P}_j \in \mathbb{R}^{3+r}$ be the $j^\text{th}$ column of $\mathtt{P}$. 
We can express the predicted camera pose as a linear combination of the columns of $\mathtt{P}$ via 
\begin{equation}
L(\mathcal{I})  = \mathbf{b} + \sum_{j=1}^n \alpha_j^\mathcal{I} \mathbf{P}_j  =  \begin{pmatrix} \hat{\mathbf{c}}_\mathcal{I} \\ \hat{\mathbf{r}}_\mathcal{I}\end{pmatrix} \enspace .
\end{equation}
We further decompose the $j^\text{th}$ column $\mathbf{P}_j$ of the projection matrix $\mathtt{P}$ into a translational part $\mathbf{c}_j \in \mathbb{R}^3$ and an orientation part $\mathbf{r}_j \in \mathbb{R}^r$, such that  $\mathbf{P}_j = (\mathbf{c}_j^T, \mathbf{r}_j^T )^T$. 
Similarly, we can decompose the bias term $\mathbf{b}$ as $\mathbf{b} = (\mathbf{c}_b^T, \mathbf{r}_b^T )^T$, resulting in 
\begin{equation}
\begin{pmatrix} \hat{\mathbf{c}}_\mathcal{I}\\ \hat{\mathbf{r}}_\mathcal{I} \end{pmatrix} = \begin{pmatrix}\mathbf{c}_b + \sum_{j=1}^n \alpha_j^\mathcal{I} \mathbf{c}_j \\ \mathbf{r}_b + \sum_{j=1}^n \alpha_j^\mathcal{I} \mathbf{r}_j \end{pmatrix} \enspace .
\label{eq:linear_pose_combination}
\end{equation}
Note that Eq.~\ref{eq:linear_pose_combination} also covers separate embeddings and projections for the position and orientation of the camera, \eg, as in~\cite{Wu2017ICRA}. 
In this case, the projection matrix has the form
\begin{equation}
\mathtt{P} =  \begin{pmatrix}\mathbf{c}_1 & \dots & \mathbf{c}_k & \mathbf{0} & \dots & \mathbf{0} \\ \mathbf{0} & \dots & \mathbf{0} & \mathbf{r}_{k+1} & \dots & \mathbf{r}_n \end{pmatrix}\enspace .
\end{equation}

\PAR{Intuitive interpretation.} Eq.~\ref{eq:linear_pose_combination} leads to the following {interpretation of  absolute camera pose regression algorithms}: 
Method such as PoseNet and its variants learn a set $\mathcal{B} = \{(\mathbf{c}_j, \mathbf{r}_j)\}$ of \emph{base poses} such that the poses of all training images can be expressed as a \emph{linear combination} of these base poses\footnote{In practice, most methods usually compute a \emph{conical} combination as they use a ReLU activation before the linear projection.}. 
How much a base pose contributes to a predicted pose depends on the appearance of the input image: 
The first stage $F(\mathcal{I})$ provides a set of feature response maps. 
The second stage $E(F(\mathcal{I}))$ then generates a high-dimensional vector $\mathbf{\alpha}^\mathcal{I}=(\alpha^\mathcal{I}_1, \dots, \alpha^\mathcal{I}_n)^T$. 
Each entry $\alpha^\mathcal{I}_j$ is computed by correlating feature activations from the first stage~\cite{Walch2017ICCV} and corresponds to a base pose $(\mathbf{c}_j, \mathbf{r}_j)$. 
These correlations thus provide the importance of each base pose for a given input image.

Fig.~\ref{fig:example_bases} visualizes the translational part $\{\mathbf{c}_j\}$ of the base poses learned by PoseNet~\cite{Kendall2015ICCV,Kendall2017CVPR} and MapNet~\cite{Brahmbhatt2018CVPR}, together with the combinations used for individual training images. 
As can be seen from the scale of the plots (in meters), $\{\mathbf{c}_j\}$ corresponds to a set of translations with small magnitude. 
Essentially, the network learns to sum these translations up to an absolute pose by scaling them appropriately via the coefficients in the embedding (\cf  Eq.~\ref{eq:linear_pose_combination}). %
For this reason, we refer to the $\{\mathbf{c}_j\}$ as \emph{base translations} rather than \emph{base positions}.
Notice that the base translations in  Fig.~\ref{fig:example_bases} approximately lie in a plane since all training poses lie in a plane. 
The supp. video (\cf Sec.~\ref{sec:appendix:video}) shows how the base translations change with changing image content.

\section{Comparison with Structure-based Methods}
\label{sec:generalization}
Visual localization algorithms represent a mapping from image content to the camera pose from which the image was taken. 
The current gold standard for localization are structure-based approaches~\cite{Sattler2017PAMI,Svarm2017PAMI,Brachmann2017CVPR,Brachmann2018CVPR,Taira2018CVPR,Schoenberger2018CVPR,Cavallari2017CVPR,Meng2017IROS}. 
These methods establish correspondences between 2D pixel positions in an image and 3D point coordinates in the scene. 
The camera pose is then computed by solving the PnP problem, \ie, by finding a pose that maximizes the number of 3D points projecting close to their corresponding 2D positions. %
As long as there are sufficiently many correct matches, structure-based methods will be able to estimate a pose. 

In contrast to structure-based methods, pose regression algorithms do not explicitly use knowledge about projective geometry. %
Rather, they learn the mapping from image content to camera pose from data. 
Based on our theory, absolute pose regression methods are expressive enough to be able to learn this mapping given enough training data:  
Changes in image content lead to different features maps $F(\mathcal{I})$, which lead to a change in the embedding $E(F(\mathcal{I}))$, and thus a different pose (\cf  Eq.~\ref{eq:linear_pose_combination}).  
Assuming the right network architecture, loss function, and enough training data, it should thus be possible to train an absolute pose regression approach that is able to accurate estimate camera poses from novel viewpoints. 

In practice, collecting a vast amount of images, computing the training poses (\eg, via SfM), and training a CNN on large amounts of data are all highly time-consuming tasks. 
Thus, methods that are able to accurately predict poses using as little training data as possible are preferable. 
In the following, we use our theoretical model to predict failure cases for pose regression techniques in scenarios with limited training data. 
We validate our predictions experimentally, thus also validating our theory. 
In addition, we show that structure-based methods, as can be expected, are able to handle these situations. 

\PAR{Experimental setup.} 
For the practical experiments used in this section, we recorded new datasets\footnote{The datasets are available at \small{\url{https://github.com/tsattler/understanding_apr}}.} . 
We deliberately limited the amount of training data to one or a few trajectories per scene and captured test images from differing viewpoints. 
Ground truth poses for training and testing data were obtained using SfM~\cite{Schoenberger2016CVPR}. 
We scaled the resulting 3D models to meters by manually measuring distances. 
For evaluation, we use both PoseNet~\cite{Kendall2015ICCV,Kendall2017CVPR} and MapNet~\cite{Brahmbhatt2018CVPR}. 
We use the PoseNet variant that learns the coefficient weighting position and orientation errors during training~\cite{Kendall2017CVPR}.
Both methods are state-of-the-art absolute pose regression algorithms. %
We use Active Search~\cite{Sattler2017PAMI} to obtain baseline results for structure-based methods. 
Active Search uses RootSIFT~\cite{Lowe04IJCV,Arandjelovic2012CVPR} features to establish 2D-3D matches. 
It is based on prioritized matching, terminating correspondence search once 200 matches have been found. 
The pose is estimated via a P3P solver~\cite{Kneip2011CVPR} inside a RANSAC~\cite{Chum08PAMI} loop, followed by non-linear refinement of the pose~\cite{ceres-solver}. 
The 3D model required by Active Search is build by matching each training image against nearby training images and triangulating the resulting matches using the provided training poses. 

\PAR{Training data captured on a line or parallel lines.} 
Let $\mathcal{T} = \{(\mathcal{I}, \mathbf{p}_\mathcal{I} = ({\mathbf{c}}_\mathcal{I}, {\mathbf{r}}_\mathcal{I}))\}$ be a set of training images with their corresponding camera poses. 
As shown in Sec.~\ref{eq:linear_pose_combination}, camera pose regression techniques express the camera pose of an image as a linear combination of a set of learned base poses. %
Consider a scenario where all training camera positions lie on a line. 
This represents the most simple and basic data capture scenario, \eg, for data captured from a car such as large-scale the San Francisco dataset~\cite{Chen2011CVPR}. 

In this scenario, each camera position ${\mathbf{c}}_\mathcal{I}$ corresponds to a point  on a line $\mathbf{o} + \delta \mathbf{d}$. 
Here, $\mathbf{o}\in \mathbb{R}^3$ is a point on the line, $\mathbf{d} \in \mathbb{R}^3$ is the direction of the line, and $\delta \in \mathbb{R}$ is a scaling factor.
One admissible solution to the training problem, although not the only one, is thus to place all base translations $\mathbf{c}_j$ on the line $\mathbf{o} + \delta \mathbf{d}$. 
As any linear combinations of points on a line lies on the line, this solution \emph{will never generalize}. %

\begin{figure*}[t!]
\begin{center}
\includegraphics[height=4.5cm]{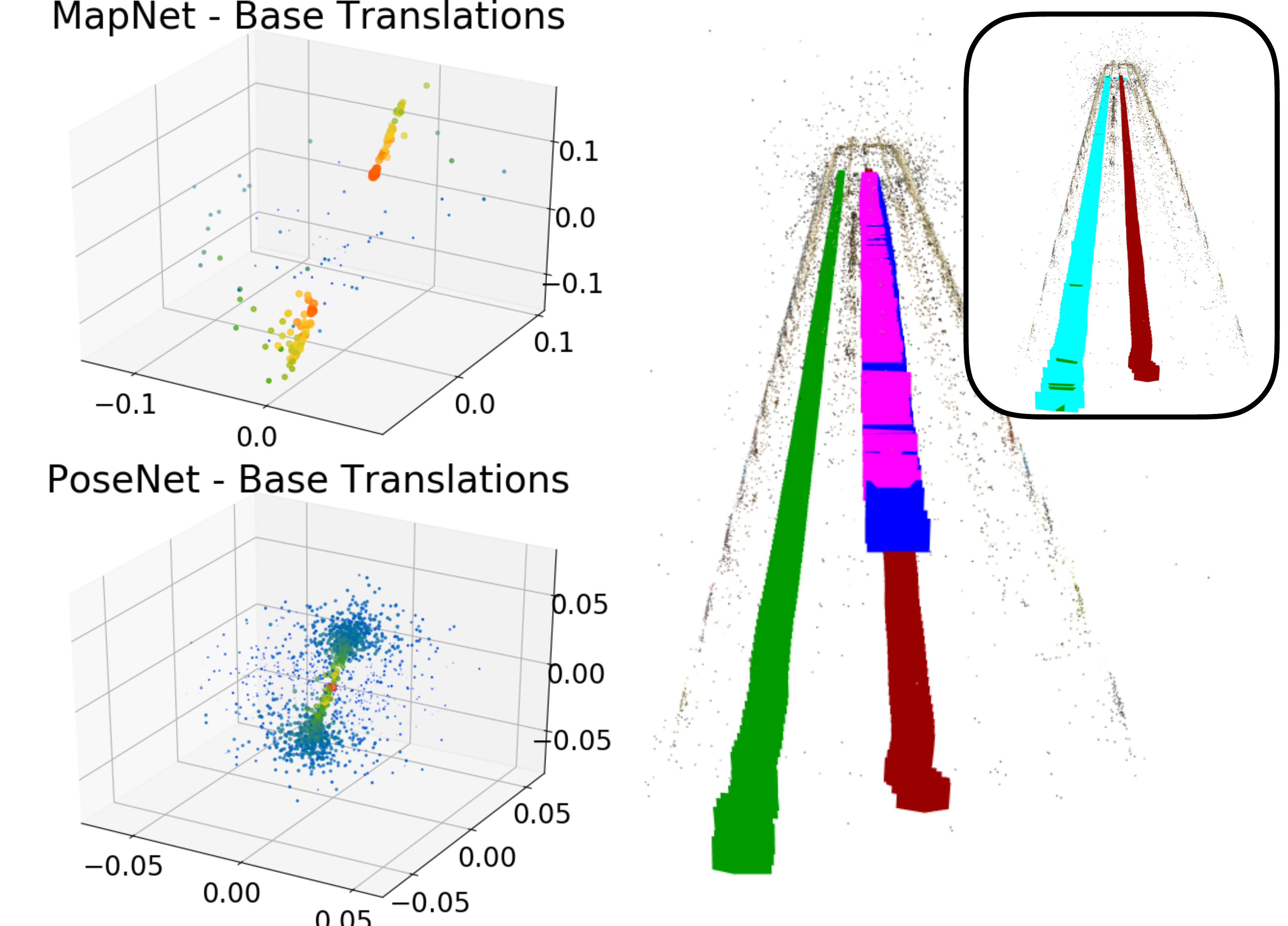}%
\hspace{2pt}%
\includegraphics[height=4.5cm]{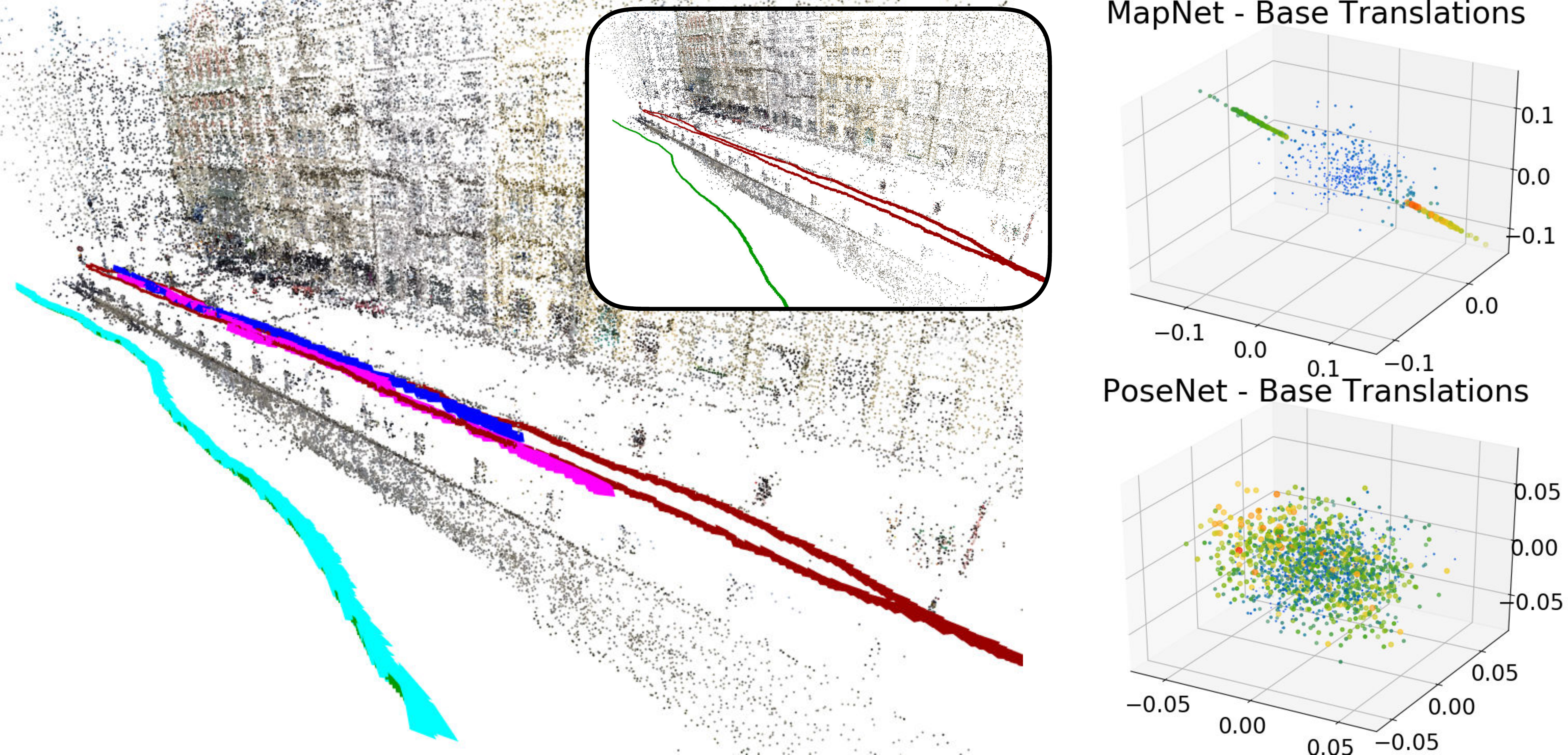}%
\end{center}
\vspace{-6pt}
\caption{Example scenarios in which two absolute pose regression techniques,  PoseNet~\cite{Kendall2015ICCV,Kendall2017CVPR} and MapNet~\cite{Brahmbhatt2018CVPR}, fail to generalize. For both scenes, the networks learn to (roughly) interpolate along a line. Consequently, the poses of the test images are also placed along this line. Please see the caption of Fig.~\ref{fig:example_bases} for details on the color coding and the base translations shown for the two scenes.}%
\label{fig:lin_motion}
\end{figure*}

Fig.~\ref{fig:lin_motion} shows two examples for this scenario: 
In the first one, training data was captured while riding an escalator upwards. 
Testing data was acquired while riding the escalator down (looking again upwards) in another lane. 
In the second example, training data was acquired while walking parallel to building facades while test data was acquired from a bit farther away. 
In both cases, MapNet clearly places most base translations along a line. %
While there are some translations not on the line, these are mostly used to handle camera shake (\cf the supp. video). 
As a result, MapNet places its estimates of the test poses on or close to the resulting line and does not generalize to diverging viewpoints. 
This clearly shows that solutions to the training problem that are guaranteed to not generalize are not only of theoretical interest but can be observed in practice. 
The base translations estimated by PoseNet are significantly more noisy and do not all lie on a line. 
Interestingly, PoseNet still places all test poses on the line corresponding to the training images. %
This shows that while the base poses span a space larger than positions on the line, PoseNet is still not able to generalize. 
This is due to a failure of mapping the image appearance to suitable weights for the base poses, showing that multiple solutions exists that do not generalize. %
As shown in Fig.~\ref{fig:lin_motion}, Active Search is able to handle both scenarios well. 

\begin{figure*}[t!]
\begin{center}
\includegraphics[height=4cm]{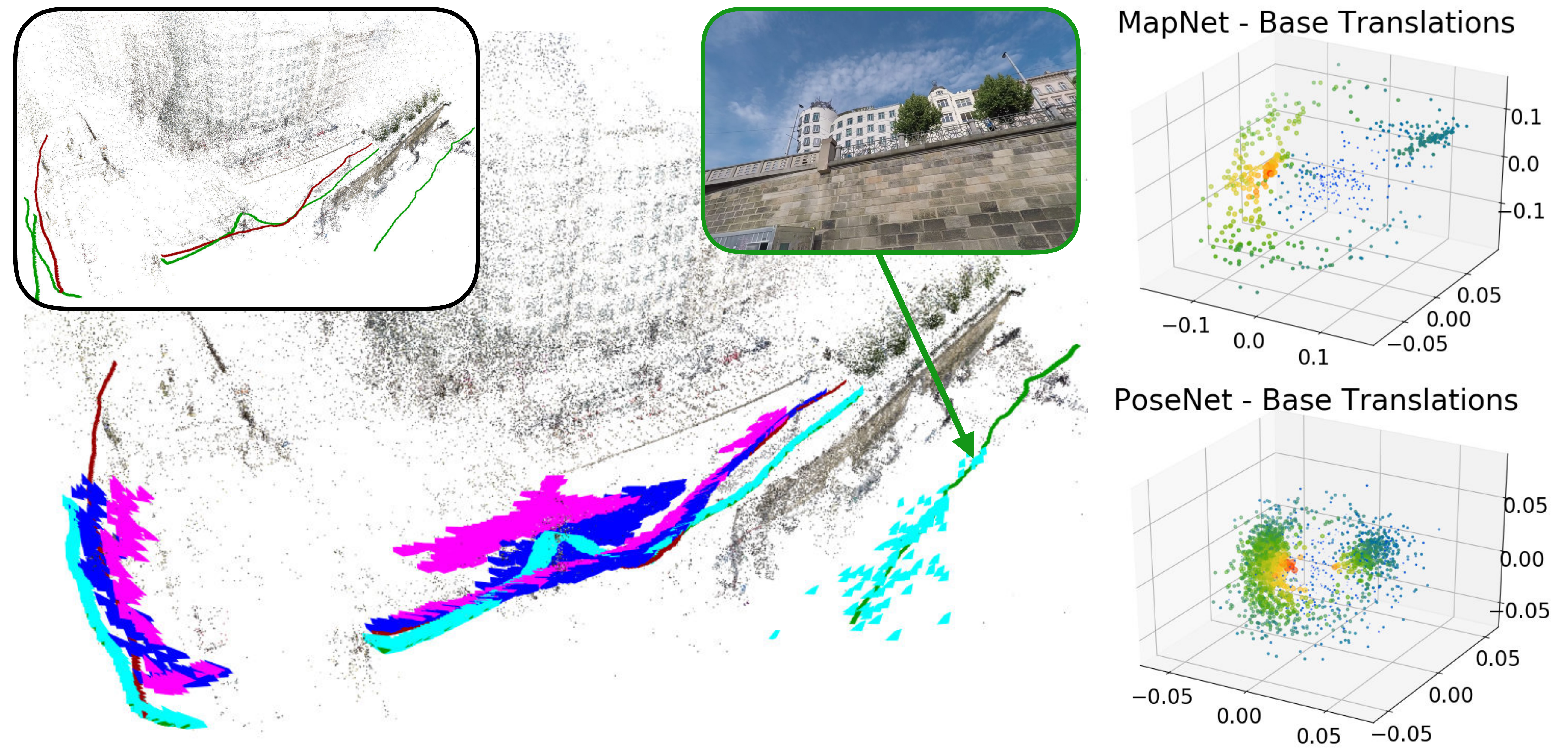}%
\includegraphics[height=4cm]{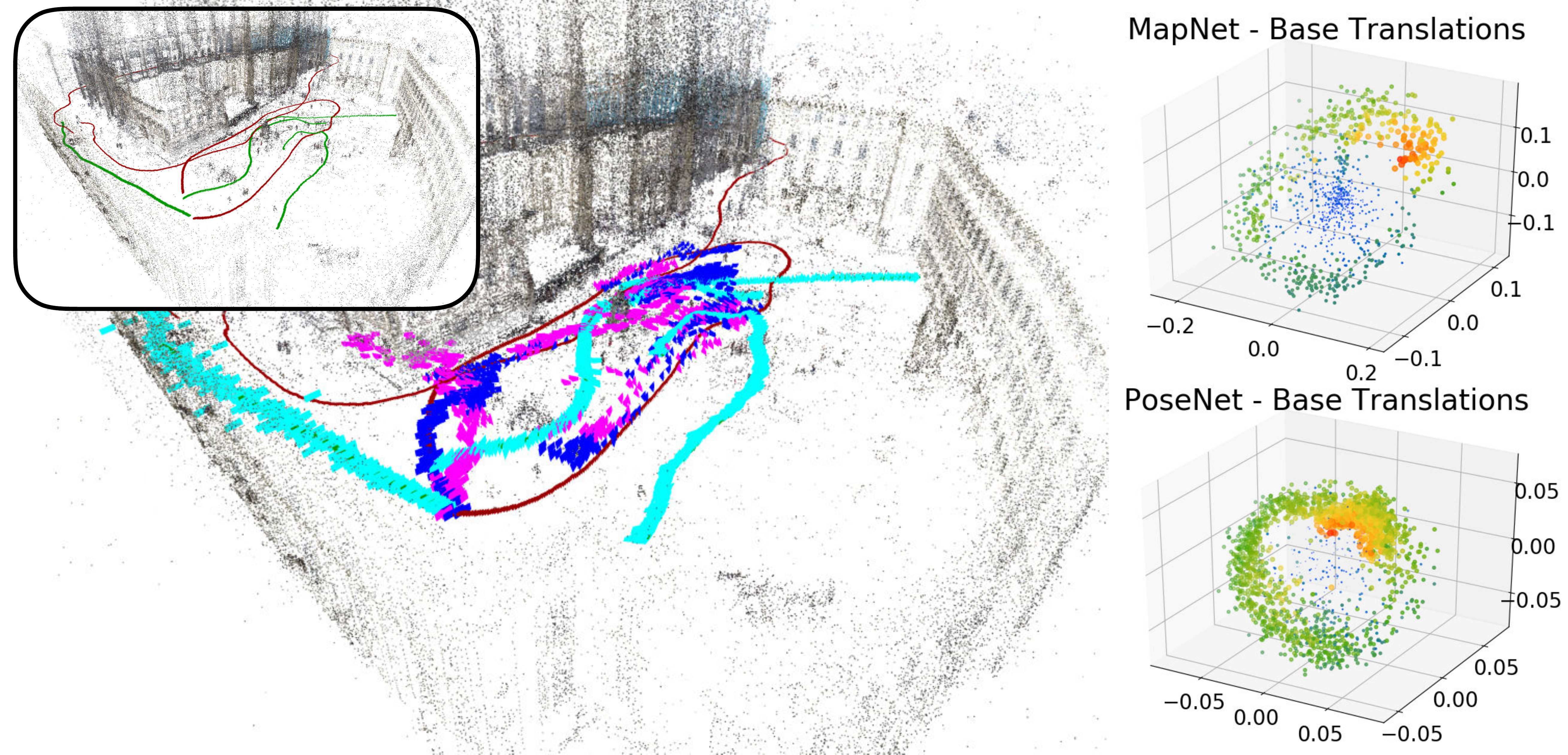}%
\end{center}
\vspace{-6pt}
\caption{Example scenarios with more general training trajectories in which two absolute pose regression techniques,  PoseNet~\cite{Kendall2015ICCV,Kendall2017CVPR} and MapNet~\cite{Brahmbhatt2018CVPR}, fail to generalize. Please see the caption of Fig.~\ref{fig:example_bases} for details on the color coding.}%
\label{fig:gen_motion}
\end{figure*}

\PAR{More general trajectories.}
The argument above exploits that it is not necessary that the base translations span the space of all possible translations to explain a set of images taken on a line. 
If the training trajectory is more general, \eg, covering all directions in a plane in the case of planar motion, this argument is not applicable anymore. 

For more general training trajectories, it is usually possible to express each viable test pose as a linear combination of the base poses. %
However, this is only a necessary but not a sufficient condition for generalization. 
As evident from Eq.~\ref{eq:linear_pose_combination}, absolute pose regression techniques couple base poses to image appearance via the coefficients $\alpha^\mathcal{I}_j$. %

Consider a part $\mathcal{P}'$ of the scene defined by a subset $\mathcal{T}' = \{\mathcal{I}\}$ of the training images. 
The corresponding relevant subset $\mathcal{B}'(\mathcal{P}')$ of the base poses $\mathcal{B} = \{(\mathbf{c}_j, \mathbf{r}_j)\}$ is %
\begin{equation}
\mathcal{B}'(\mathcal{P}') =  \{(\mathbf{c}_j, \mathbf{r}_j) | \text{ exists } \mathcal{I} \in \mathcal{T}' \text{ with } |\alpha^\mathcal{I}_j| > 0\} \enspace .
\end{equation}
A stronger necessary condition for generalization is that the linear span of each such $\mathcal{B}'(\mathcal{P}')$ contains the poses of all test images in $\mathcal{P}'$\footnote{This condition is not sufficient as a network might not learn the "right" embedding for expressing all test poses as linear combinations of $\mathcal{B}'(\mathcal{P}')$.}. 
In the following, we show that this is not necessarily guaranteed in practice. 

Figs.~\ref{fig:example_bases} and \ref{fig:gen_motion} show scenes with more general motion. 
For each scene, we show the training and ground truth testing trajectories, as well as the test trajectories estimated by PoseNet, MapNet, and Active Search. 
In addition, we show the base translations used by the two networks. 
Since the training images are taken in a plane, the base translations also lie in a plane (up to some noise). 
As can be seen, the networks are able to generalize in some parts of the scene, \eg, when the test trajectory crosses the training trajectory in Fig.~\ref{fig:example_bases}. 
In other parts, they however seem to resort to some form of nearest neighbor strategy: 
Test poses are placed close to parts of the training trajectory with similar image appearance. 
In these parts, the relevant base translations are not sufficient to model the test positions more accurately. %
This shows that more training data is required in these regions.
It also shows that networks do not automatically benefit from recording more data in unrelated parts of the scene. %

As can be expected, Active Search fails or produces inaccurate pose estimates when there is little visual overlap between the test and training images (\cf the example test image in  Fig.~\ref{fig:gen_motion}(left), where the wall visible in the image is not seen during training). 
Still, Active Search overall handles viewpoint changes significantly better. %

\begin{figure}[t!]
\begin{center}
\includegraphics[height=4cm]{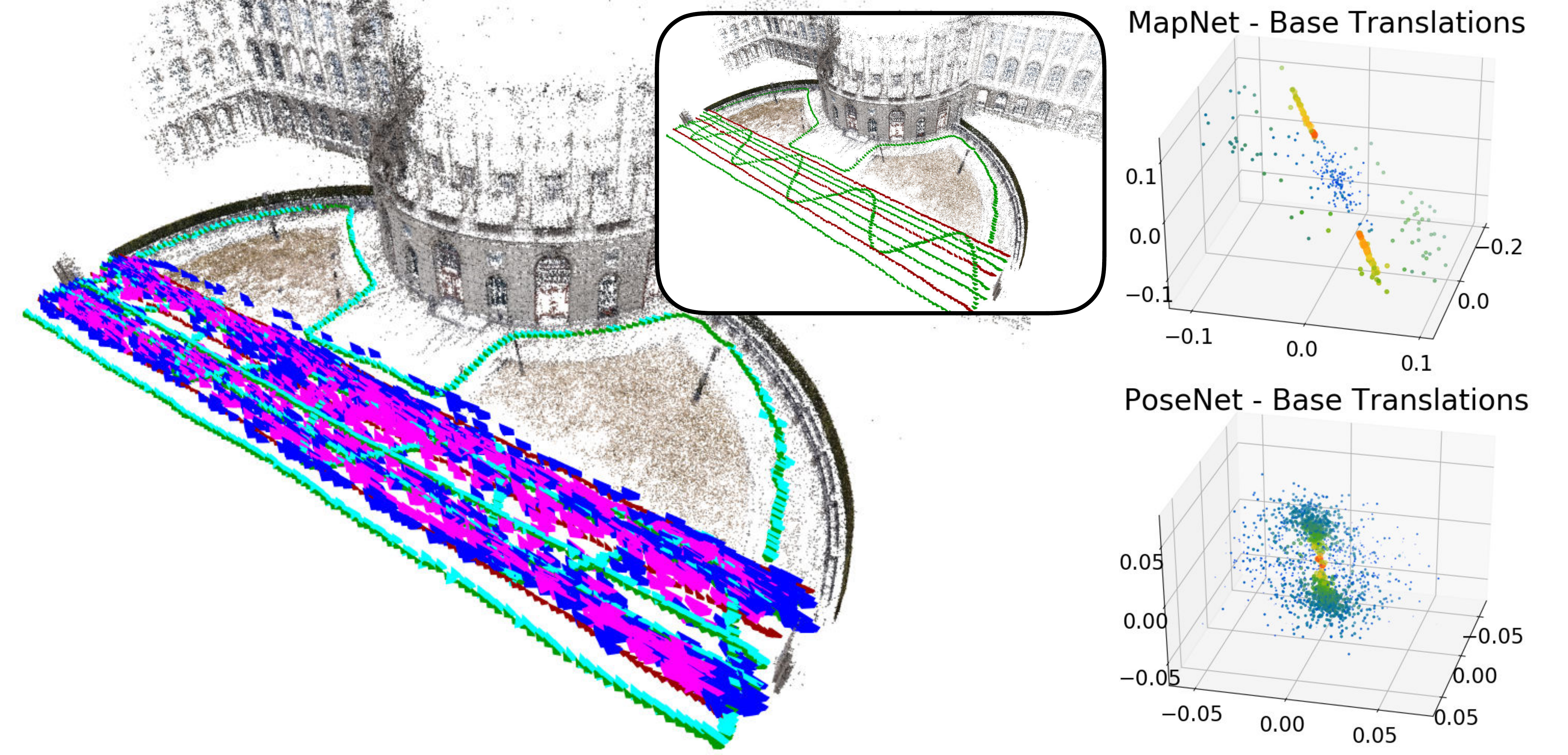}%
\end{center}
\vspace{-6pt}
\caption{See caption of Fig.~\ref{fig:example_bases} for details.}
\label{fig:gen_motion2}
\end{figure}

Fig.~\ref{fig:gen_motion2} shows a more complex example, where the training data is captured on multiple parallel lines and should be sufficient to explain the test poses.  
In this case, both networks are able to estimate poses close to these lines, but are not able to properly interpolate between them and do not generalize beyond them. 
Active Search is mostly handles the large viewpoint changes between training and testing images. 
If the change is too large however, it fails to find enough matches and thus to estimate a pose. %
Local features that are more robust to large viewpoint changes are an active field of research~\cite{Ono2018ARXIV,Rocco2018NIPS} and structure-based methods will automatically benefit from progress in this field. %

\begin{table*}
\scriptsize{
\setlength{\tabcolsep}{2pt}
\begin{center}
\begin{tabular}{l|c|c|c|c|c|c|c|c|c|c}
\multicolumn{1}{r|}{max. distance} & - & \multicolumn{3}{c|}{1m} &   \multicolumn{3}{c|}{2m} &   \multicolumn{3}{c}{3m} \\
\multicolumn{1}{r|}{spacing} & - & 1m & 0.5m & 0.25m & 1m & 0.5m & 0.25m &  1m & 0.5m & 0.25m \\
\multicolumn{1}{r|}{\# training images} & 203 & 501 & 1,315 & 4,576 & 683 & 2,035 & 7,425 &  806 & 2,531 & 9,412 \\ \hline
PoseNet~\cite{Kendall2017CVPR} & 1.19 / 6.88 & 1.02 / 6.48 & 0.74 / 7.07  & 0.79 / 5.84 & 1.15 / 8.10 & 0.86 / 6.88 & 0.54 / 5.84 & 0.66 / 6.88 & 0.66 / 6.06 & 0.68 / 5.38 \\ \hline
MapNet~\cite{Brahmbhatt2018CVPR} &  1.07 / 4.70  & 0.61 / 3.31 & 0.64 / 2.85 &   0.41 / 2.18  & 0.72 / 3.41 & 0.42 / 2.06 &  0.38 / 2.31  &  0.69 / 3.18 &  0.44 / 2.39  &  0.33 / 1.46   \\ \hline 
 Active Search~\cite{Sattler2017PAMI} & \textbf{0.01} / \textbf{0.04}  \\ \hline 
DenseVLAD~\cite{Torii15CVPR} & 0.98 / 7.90 & 0.79 / 8.01 & 0.74 / 7.81 & 0.63 / 7.68 & 0.72 / 7.81 & 0.61 / 7.38 & 0.57 / 6.94 & 0.66 / 7.81 & 0.60 / 7.27 & 0.51 / 6.87 \\
DenseVLAD+Inter. & 0.89 / 5.71 & 0.75 / 5.62 & 0.52 / 6.65 & 0.45 / 6.93 & 0.57 / 5.96 & 0.48 / 6.13 & 0.41 / 6.41 & 0.49 / 6.07 & 0.46 / 6.26 & 0.38 / 6.41\\ \hline 
\end{tabular}%
\end{center}%
}%
\vspace{-6pt}
\caption{Median position / orientation errors in meters / degree on the \textbf{synthetic Shop Facade} dataset obtained by rendering a multi-view stereo reconstruction. We enhance the training set by additional images captured on a regular grid, varying the spacing between images. We only consider additional images within a certain maximum distance to the positions of the original training poses.}
\label{tab:experiments:synthetic}
\end{table*}

\PAR{Using densely sampled training data.} 
Training using more data in a part of the scene should intuitively improve the prediction accuracy of pose regression techniques. 
To verify this assumption, we use synthetic data: 
We created a 3D model of the Shop Facade scenes from the Cambridge Landmarks dataset~\cite{Kendall2015ICCV} using multi-view stereo~\cite{Schoenberger2016ECCV}. 
We then rendered~\cite{Waechter2017SIGGRAPH} the scene from the poses of the original training and testing images, as well as from a set of additional poses. 
These poses are placed on a regular grid in the plane containing the original poses, with a spacing of 25cm between poses. 
We only created poses up to 3 meters away from the original training poses. 
The orientation of each additional pose is set to that of the nearest training pose.
Varying the maximum distance to the original poses and the grid spacing thus creates varying amounts of training data.

Tab.~\ref{tab:experiments:synthetic} compares PoseNet and MapNet trained on varying amounts of data with Active Search using only renderings from the original training poses\footnote{All images used in this experiment are renderings of the 3D model. We use a resolution of 455$\times$256 pixels as input to all methods.}. 
As expected, using more training data improves pose accuracy. %
However, PoseNet and MapNet do not perform even close to Active Search, even with one order of magnitude more data. %

\PAR{Discussion.}
Pose regression techniques are unlikely to work well when only little training data is available and significant viewpoint changes need to be handled. %
This clearly limits their relevance for practical applications. 
Even with large amounts of training data, pose regression does not reach the same performance as structure-based methods. 
This clearly shows a fundamental conceptual difference between the two approaches to visual localization. 
We attribute this divide to the fact that the latter are based on the laws of projective geometry and the underlying 3D geometry of the scene. 

\section{Comparison with Image Retrieval}
\label{sec:practical_experiments}
As can be seen in Fig.~\ref{fig:example_bases} and Fig.~\ref{fig:gen_motion}(right), absolute pose regression (APR) techniques tend to predict test poses close to the training poses in regions where little training data is available. 
This behavior is similar to that of image retrieval approaches. 
Below, we show %
that this behavioral similarity is not a coincident. 
Rather, there is a strong connection between APR and image retrieval. 
We also show that APR methods do not consistently outperform a  retrieval baseline. 

\PAR{Relation to image retrieval.}
Let $\mathcal{I}$ be a test image and $\mathcal{J}$ a training image observing the same part of the scene. 
We can write the embedding $\alpha^\mathcal{I}$ as $\alpha^\mathcal{I} = \alpha^\mathcal{J} + \Delta^{\mathcal{I}}$, for some offset $\Delta^{\mathcal{I}}$. 
Using Eq.~\ref{eq:linear_pose_combination}, we can thus relate the pose $(\hat{\mathbf{c}}_\mathcal{I}, \hat{\mathbf{r}}_\mathcal{I})$ estimated for $\mathcal{I}$ to the pose $(\hat{\mathbf{c}}_\mathcal{J}, \hat{\mathbf{r}}_\mathcal{J})$ estimated for $\mathcal{J}$ via
\begin{equation}
\begin{pmatrix} \hat{\mathbf{c}}_\mathcal{I}\\ \hat{\mathbf{r}}_\mathcal{I} \end{pmatrix} \! = \! \begin{pmatrix} \hat{\mathbf{c}}_\mathcal{J}\\ \hat{\mathbf{r}}_\mathcal{J} \end{pmatrix} \! + \! \begin{pmatrix} \sum_{j=1}^n \Delta^{\mathcal{I}}_j \mathbf{c}_j \\ \sum_{j=1}^n \Delta^{\mathcal{I}}_j \mathbf{r}_j\end{pmatrix} \! = \! \begin{pmatrix} \hat{\mathbf{c}}_\mathcal{J}\\ \hat{\mathbf{r}}_\mathcal{J} \end{pmatrix} \! + \! \begin{pmatrix} \hat{\mathbf{c}}_{\mathcal{I},\mathcal{J}}\\ \hat{\mathbf{r}}_{\mathcal{I},\mathcal{J}} \end{pmatrix}  \! .
\label{eq:retrieval_plus_offset}
\end{equation}
Here, $(\hat{\mathbf{c}}_{\mathcal{I},\mathcal{J}}, \hat{\mathbf{r}}_{\mathcal{I},\mathcal{J}})$ is a pose offset, \ie, the pose of $\mathcal{I}$ is predicted relative to the pose predicted for $\mathcal{J}$. %

Eq.~\ref{eq:retrieval_plus_offset} highlights the conceptual similarity between absolute  pose regression and image retrieval. 
Standard image retrieval approaches first find the training image $\mathcal{J}$ most similar to a given test image $\mathcal{I}$, where similarity is defined in some feature space such as Bag-of-Words (BoW)~\cite{Sivic03ICCV} or VLAD~\cite{Torii15CVPR,Arandjelovic16CVPR,Jegou2012PAMI}. 
The pose of the test image is then approximated via the pose of the retrieved image, \ie, 
$(\hat{\mathbf{c}}_\mathcal{I}, \hat{\mathbf{r}}_\mathcal{I} )  =  ({\mathbf{c}}_\mathcal{J}, {\mathbf{r}}_\mathcal{J})$, without adding an offset.  
However, retrieval methods can also estimate such an offset %
 as an affine combination $\sum_{i=1}^k a_i ({\mathbf{c}}_{\mathcal{J}_i}, {\mathbf{r}}_{\mathcal{J}_i})$, $\sum a_i = 1$, of the poses of top-$k$ retrieved training images $\mathcal{J}_1, \dots, \mathcal{J}_k$. %
Let $\mathbf{d}(\mathcal{I})$ be the descriptor for image $\mathcal{I}$ used during retrieval. 
The weights $a_i$ can be obtained by finding the affine combination %
of training image descriptors that is closest to the test  descriptor $\mathbf{d}(\mathcal{I})$, \ie, by minimizing $|| \mathbf{d}(\mathcal{I}) - \sum_{i=1}^k a_i \mathbf{d}(\mathcal{J}_i)||_2$ subject to $\sum a_i = 1$. 
This approach has been shown to work well for linearly interpolating between two BoW representations~\cite{Torii2011ICCVW}. 
Note the conceptual similarity between this interpolation and Eq.~\ref{eq:linear_pose_combination}, where the trained base poses are used instead of the poses of the retrieved images. %

Eq.~\ref{eq:retrieval_plus_offset} also establishes a relation between APR approaches and relative pose regression (RPR) algorithms. 
RPR methods first identify a set of training images relevant to a given test image, \eg, using image retrieval~\cite{Balntas2018ECCV,Laskar2017ICCVW} or by encoding the training images in a CNN~\cite{Saha2018BMVC}. 
They then compute a pose offset from the training images to the test image via regression. 
RPR approaches naturally benefit from computing offsets to multiple training images~\cite{Balntas2018ECCV}. 

\subsection{Experimental Comparison}
\PAR{Baselines.} 
We use \emph{DenseVLAD}~\cite{Torii15CVPR} as an image retrieval baseline.  
DenseVLAD densely extracts RootSIFT~\cite{Arandjelovic2012CVPR,Lowe04IJCV} descriptors from an image and pools them into a VLAD~\cite{Jegou2012PAMI} descriptor. 
Dimensionality reduction via PCA, trained on an unrelated outdoor dataset~\cite{Torii15CVPR}, is then used to reduce the dimensionality of the descriptor to 4096. 
The Euclidean distance is used to measure similarity between two DenseVLAD descriptors. 
We use the implementation provided by~\cite{Torii15CVPR}, but only extract RootSIFT descriptors at a single scale\footnote{Scale invariance is not desirable when searching for the training image taken from the most similar pose.}. 
We chose DenseVLAD as it uses a handcrafted feature representation. 
At the same time, DenseVLAD has been shown to perform well even on challenging localization tasks~\cite{Torii15CVPR,Sattler2017CVPR,Sattler2018CVPR}. 
However, DenseVLAD has not yet been used as a baseline for pose regression.

DenseVLAD approximates the pose of the test image via the pose of the most similar training image. 
In addition, we also use 
a variant, denoted as  \emph{DenseVLAD + Inter.}, that uses the interpolation approach described above. 
We use all top-$k$ ranked images for interpolation. 
As there might be some outliers among the top retrieved images, interpolation can potentially decrease pose accuracy. %
However, we decided to keep this baseline as simple as possible and thus did not implement an outlier filtering mechanism.

\PAR{Cambridge Landmarks~\cite{Kendall2015ICCV} and 7 Scenes~\cite{Shotton2013CVPR}.} 
In a %
first experiment, we compare state-of-the-art pose regression techniques to the two image retrieval baselines on the Cambridge Landmarks~\cite{Kendall2015ICCV} and 7 Scenes~\cite{Shotton2013CVPR} datasets. 
These two relatively small-scale datasets are commonly used to evaluate pose regression approaches. 
We only compare methods that predict a camera pose from a single image. 

\begin{table*}
\scriptsize{
\setlength{\tabcolsep}{3pt}
\begin{center}
\begin{tabular}{cl|c|c|c|c|c||c|c|c|c|c|c|c}
& & \multicolumn{5}{c||}{Cambridge Landmarks} & \multicolumn{7}{c}{7 Scenes} \\
 & & Kings & Old & Shop & St. Mary's & Street & Chess & Fire & Heads & Office & Pumpkin & Kitchen & Stairs  \\ \hline
\multirow{12}*{\begin{sideways} APR \end{sideways}}& PoseNet (PN)~\cite{Kendall2015ICCV} & \textcolor{red}{1.92}/\textcolor{red}{5.40} & 2.31/\textcolor{red}{5.38} & \textcolor{red}{1.46}/\textcolor{red}{8.08} & \textcolor{red}{2.65}/\textcolor{red}{8.48} & & \textcolor{red}{0.32}/8.12 & \textcolor{red}{0.47}/\textcolor{red}{14.4} & \textcolor{red}{0.29}/12.0 & \textcolor{red}{0.48}/7.68 & \textcolor{red}{0.47}/8.42 & \textcolor{red}{0.59}/8.64 & \textcolor{red}{0.47}/13.8 \\

& PN learned weights~\cite{Kendall2017CVPR} & 0.99/1.06 & 2.17/2.94 & \textcolor{red}{1.05}/3.97 & 1.49/3.43 & \textcolor{red}{20.7}/\textcolor{red}{25.7} & 0.14/4.50 & 0.27/11.8 & \textcolor{red}{0.18}/12.1 & 0.20/5.77 & 0.25/4.82 & 0.24/5.52 & \textcolor{red}{0.37}/10.6\\

& Bay. PN~\cite{Kendall2016ICRA} & \textcolor{red}{1.74}/4.06 & 2.57/\textcolor{red}{5.14} & \textcolor{red}{1.25}/\textcolor{red}{7.54} & \textcolor{red}{2.11}/\textcolor{red}{8.38} & & \textcolor{red}{0.37}/7.24 & \textcolor{red}{0.43}/\textcolor{red}{13.7} & \textcolor{red}{0.31}/12.0 & \textcolor{red}{0.48}/8.04 & \textcolor{red}{0.61}/7.08 & \textcolor{red}{0.58}/7.54 & \textcolor{red}{0.48}/13.1\\

& geo. PN~\cite{Kendall2017CVPR} & 0.88/1.04 & \textcolor{red}{3.20}/3.29 & 0.88/3.78 & 1.57/3.32 & \textcolor{red}{20.3}/\textcolor{red}{25.5} & 0.13/4.48 & 0.27/11.3 & \textcolor{red}{0.17}/13.0 & 0.19/5.55 & \textcolor{red}{0.26}/4.75 & 0.23/5.35 & \textcolor{red}{0.35}/12.4\\

& LSTM PN~\cite{Walch2017ICCV} & 0.99/3.65 & 1.51/4.29 & \textcolor{red}{1.18}/\textcolor{red}{7.44} & 1.52/\textcolor{red}{6.68} & & \textcolor{red}{0.24}/5.77 & \textcolor{red}{0.34}/11.9 & \textcolor{red}{0.21}/13.7 & \textcolor{red}{0.30}/8.08 & \textcolor{red}{0.33}/7.00 & \textcolor{red}{0.37}/8.83 & \textcolor{red}{0.40}/13.7\\

& GPoseNet~\cite{Cai2018BMVC} & \textcolor{red}{1.61}/2.29 & 2.62/3.89 & \textcolor{red}{1.14}/5.73 & \textcolor{red}{2.93}/6.46 & & \textcolor{red}{0.20}/7.11  & \textcolor{red}{0.38}/12.3 & \textcolor{red}{0.21}/13.8 & \textcolor{red}{0.28}/ 8.83 & \textcolor{red}{0.37}/6.94 & \textcolor{red}{0.35}/8.15 & \textcolor{red}{0.37}/12.5\\

& SVS-Pose~\cite{Naseer2017IROS} & 1.06/2.81 & 1.50/4.03 & 0.63/\textcolor{red}{5.73} & \textcolor{red}{2.11}/\textcolor{red}{8.11} & & & & & & & \\

 & Hourglass PN~\cite{Melekhov2017ICCVW} & & & & & & 0.15/6.17 & 0.27/10.8 & \textcolor{red}{0.19}/11.6 & 0.21/8.48 & 0.25/7.01 & \textcolor{red}{0.27}/10.2 & \textcolor{red}{0.29}/12.5\\
 
 & BranchNet~\cite{Wu2017ICRA} & & & & & & \textcolor{red}{0.18}/5.17 & \textcolor{red}{0.34}/8.99 & \textcolor{red}{0.20}/14.2 & \textcolor{red}{0.30}/7.05 & \textcolor{red}{0.27}/5.10 & \textcolor{red}{0.33}/7.40 & \textcolor{red}{0.38}/10.3\\
 
& MapNet~\cite{Brahmbhatt2018CVPR} & 1.07/1.89 & 1.94/3.91 & \textcolor{red}{1.49}/4.22 & \textcolor{red}{2.00}/4.53 & & 0.08/3.25 & 0.27/11.7 & \textcolor{red}{0.18}/13.3 & 0.17/5.15 & 0.22/4.02 & 0.23/4.93 &  \textcolor{red}{0.30}/12.1  \\

& MapNet+~\cite{Brahmbhatt2018CVPR} & & & & & & 0.10/3.17 & 0.20/9.04 &  0.13/11.1 & 0.18/5.38 & 0.19/3.92 & 0.20/5.01 &  \textcolor{red}{0.30}/13.4 \\ 

& MapNet+PGO~\cite{Brahmbhatt2018CVPR} & & & & & & 0.09/3.24 & 0.20/9.29 &  0.12/8.45 & 0.19/5.42 & 0.19/3.96 & 0.20/4.94 &  \textcolor{red}{0.27}/10.6 \\ \hline

\multirow{5}*{\begin{sideways} RPR \end{sideways}} & Relative PN~\cite{Laskar2017ICCVW} (U) & & & & & & \textcolor{red}{0.31}/\textcolor{red}{15.0} & \textcolor{red}{0.40}/\textcolor{red}{19.0} & \textcolor{red}{0.24}/\textcolor{red}{22.2} & \textcolor{red}{0.38}/\textcolor{red}{14.1} & \textcolor{red}{0.44}/\textcolor{red}{18.2} & \textcolor{red}{0.41}/\textcolor{red}{16.5} & \textcolor{red}{0.35}/\textcolor{red}{23.6} \\
 
  & Relative PN~\cite{Laskar2017ICCVW} (7S) & & & & & & 0.13/6.46 & 0.26/\textcolor{red}{12.7} & \textcolor{red}{0.14}/12.3 & 0.21/7.35 & 0.24/6.35 & 0.24/8.03 & \textcolor{red}{0.27}/11.8\\
  
& RelocNet~\cite{Balntas2018ECCV} (SN) & & & & &  & \textcolor{red}{0.21}/ \textcolor{red}{10.9}  &  \textcolor{black}{0.32}/11.8  &  \textcolor{red}{0.15}/13.4  &  \textcolor{red}{0.31}/ \textcolor{red}{10.3}  &  \textcolor{red}{0.40}/ \textcolor{red}{10.9}  &  \textcolor{red}{0.33}/10.3 &  \textcolor{red}{0.33}/11.4\\

& RelocNet~\cite{Balntas2018ECCV} (7S) & & & & & & 0.12/4.14 & 0.26/10.4 & \textcolor{red}{0.14}/10.5 & 0.18/5.32 & \textcolor{red}{0.26}/4.17 & 0.23/5.08  &  \textcolor{red}{0.28}/7.53\\ 

& AnchorNet~\cite{Saha2018BMVC} & 0.57/0.88 & 1.21/2.55 & 0.52/2.27 & 1.04/2.69 & \textcolor{red}{7.86}/24.2 & 0.06/3.89 & 0.15/10.3 & 0.08/10.9 & 0.09/5.15 & 0.10/2.97 & 0.08/4.68 & 0.10/9.26\\ \hline \hline 

\multirow{2}*{\begin{sideways} IR \end{sideways}} & DenseVLAD~\cite{Torii15CVPR} & 2.80/5.72  & 4.01/7.13 & 1.11/7.61 & 2.31/8.00 & 5.16/23.5 & 0.21/12.5 & 0.33/13.8 & 0.15/14.9 & 0.28/11.2 & 0.31/11.3 & 0.30/12.3 & 0.25/15.8 \\

& DenseVLAD + Inter. & 1.48/4.45 & 2.68/4.63 & 0.90/4.32 & 1.62/6.06 & 15.4/25.7 & 0.18/10.0 & 0.33/12.4 & 0.14/14.3 & 0.25/10.1 & 0.26/9.42 & 0.27/11.1 & 0.24/14.7\\ \hline \hline %

\multirow{4}*{\begin{sideways} 3D \end{sideways}} & Active Search~\cite{Sattler2017PAMI} & 0.42/0.55 & 0.44/1.01 & 0.12/0.40 & 0.19/0.54 & \textbf{0.85}/\textbf{0.8} & 0.04/1.96 & 0.03/1.53 & 0.02/1.45 & 0.09/3.61 & 0.08/3.10 & 0.07/3.37 & \textbf{0.03}/\textbf{2.22}\\
& BTBRF~\cite{Meng2017IROS} & 0.39/0.36 & 0.30/0.41 & 0.15/0.31 & 0.20/\textbf{0.40} &  &  &  & & &  &  & \\
  & DSAC++~\cite{Brachmann2018CVPR} & \textbf{0.18}/\textbf{0.3} & \textbf{0.20}/\textbf{0.3} & \textbf{0.06}/\textbf{0.3} & \textbf{0.13}/\textbf{0.4} & & \textbf{0.02}/\textbf{0.5} & \textbf{0.02}/\textbf{0.9} & \textbf{0.01}/\textbf{0.8} & \textbf{0.03}/\textbf{0.7} & \textbf{0.04}/\textbf{1.1} & \textbf{0.04}/\textbf{1.1} & 0.09/2.6\\
  & InLoc~\cite{Taira2018CVPR} & & & & & & 0.03/1.05 & 0.03/1.07 & 0.02/1.16 & \textbf{0.03}/1.05 & 0.05/1.55 & \textbf{0.04}/1.31 & 0.09/2.47\\
\end{tabular}
\end{center}
}%
\vspace{-6pt}
\caption{Results on the \textbf{Cambridge Landmarks}~\cite{Kendall2015ICCV} and \textbf{7 Scenes}~\cite{Shotton2013CVPR} datasets. We compare absolute (APR) and relative (RPR) pose regression methods, image retrieval (IR) techniques, and structure-based (3D) approaches. We report the median position / orientation error in meters / degree. \emph{DenseVLAD + Inter.} uses the top-20 (Cambridge Landmarks) respectively top-25 (7 Scenes) retrieved images.  \textcolor{red}{Red} numbers show when a method fails to outperform the image retrieval (IR) baselines. Results for Cambridge Landmarks for MapNet are obtained running the code of the authors.}%
\label{tab:experiments:cambridge_and_7}
\end{table*}

Tab.~\ref{tab:experiments:cambridge_and_7} shows the median position and orientation errors obtained by the various methods. 
As can be seen by the results marked in red, \emph{none of the absolute and relative pose regression approaches is able to consistently outperform the retrieval baselines}. %
In addition, \emph{pose regression techniques are often closer in performance to image retrieval than to structure-based methods}. 
In particular, these results verify our theoretical analysis that APR is much closer related to image retrieval than to structure-based methods. %

Out of the four best-performing pose regression approaches (MapNet~\cite{Brahmbhatt2018CVPR}, RelocNet~\cite{Balntas2018ECCV}, Relative PN~\cite{Laskar2017ICCVW}, AnchorNet~\cite{Saha2018BMVC}), three are RPR approaches (RelocNet, Relative PN, AnchorNet). %
AnchorNet comes closest to structure-based methods. 
It uses a brute-force approach that essentially estimates a pose offset between the input image and every 10$^\text{th}$ training image. 
Considering the relative improvement\footnote{Defined as the ratio of the position / orientation errors of two methods.}, AnchorNet typically performs closer to other APR or RPR methods than to the best performing structure-based approach in each scene. 
It also fails to outperform the simple DenseVLAD baseline on the Street scene, which is the largest and most complex scene in the Cambridge Landmarks dataset~\cite{Brachmann2018CVPR,Naseer2017IROS}.

AnchorNet encodes the training images in the regression CNN and thus needs to be trained specifically per scene. 
In contrast, Relative PN and RelocNet 
both perform an explicit image retrieval step. 
They can thus also be trained on unrelated scenes. 
Besides RelocNet and Relative PN trained on 7 Scenes (7S), we thus also compare against variants trained on other datasets (ScanNet (SN)~\cite{Dai2017CVPR},  University (U)~\cite{Laskar2017ICCVW}). 
As shown in Tab.~\ref{tab:experiments:cambridge_and_7}, both approaches currently do not generalize well using this data, as they are less accurate than DenseVLAD (which requires no training).

One challenge of the Cambridge Landmarks and 7 Scenes datasets is that there are significant differences in pose between the training and test images. %
As shown in Sec.~\ref{sec:generalization}, this is  a severe challenge for current regression techniques. %
In the following, we focus on scenes with less deviation between training and test poses,  %
which should be much easier for pose regression techniques. 
We show results on two such datasets. 
A further experiment (on the DeepLoc dataset~\cite{Radwan2018RAL}) can be found in Sec.~\ref{sec:appendix:add_experiment}.

\begin{table}
\footnotesize{
\setlength{\tabcolsep}{2pt}
\begin{center}
\begin{tabular}{c|c|c|c|c}
PoseNet & MapNet & LSTM  & Dense & DenseVLAD \\
\cite{Kendall2015ICCV} & ~\cite{Brahmbhatt2018CVPR} & PN~\cite{Walch2017ICCV}  & VLAD~\cite{Torii15CVPR} & +Inter. \\ \hline
1.87m, 6.14$^\circ$ & 1.71m, 3.50$^\circ$ & 1.31m, 2.79$^\circ$ & 1.08m, 1.82$^\circ$ & 0.49m, 2.01$^\circ$
\end{tabular}
\end{center}
}\vspace{-6pt}
\caption{Median position and orientation errors on the \textbf{TUM LSI} dataset~\cite{Walch2017ICCV}. The top-2 retrieved images are used for interpolation.}
\label{tab:experiments:tum_lsi}
\end{table}

\PAR{TUM LSI~\cite{Walch2017ICCV}.} 
The scenes in the Cambridge Landmarks and 7 Scenes datasets are typically rather well-textured. 
Thus, we can expect that the SIFT descriptors used by DenseVLAD and Active Search~\cite{Sattler2017PAMI} work rather well. 
In contrast, the TUM LSI indoor dataset~\cite{Walch2017ICCV} contains large textureless walls and repeating structures. 
In general, we would expect learned approaches to perform significantly better than methods based on low-level SIFT features as the former can learn to use higher-level structures. %
Yet, as shown in Tab.~\ref{tab:experiments:tum_lsi}, DenseVLAD still outperforms pose regression techniques on this more challenging dataset. 

\begin{figure}
\begin{center}
\includegraphics[width=0.49\linewidth]{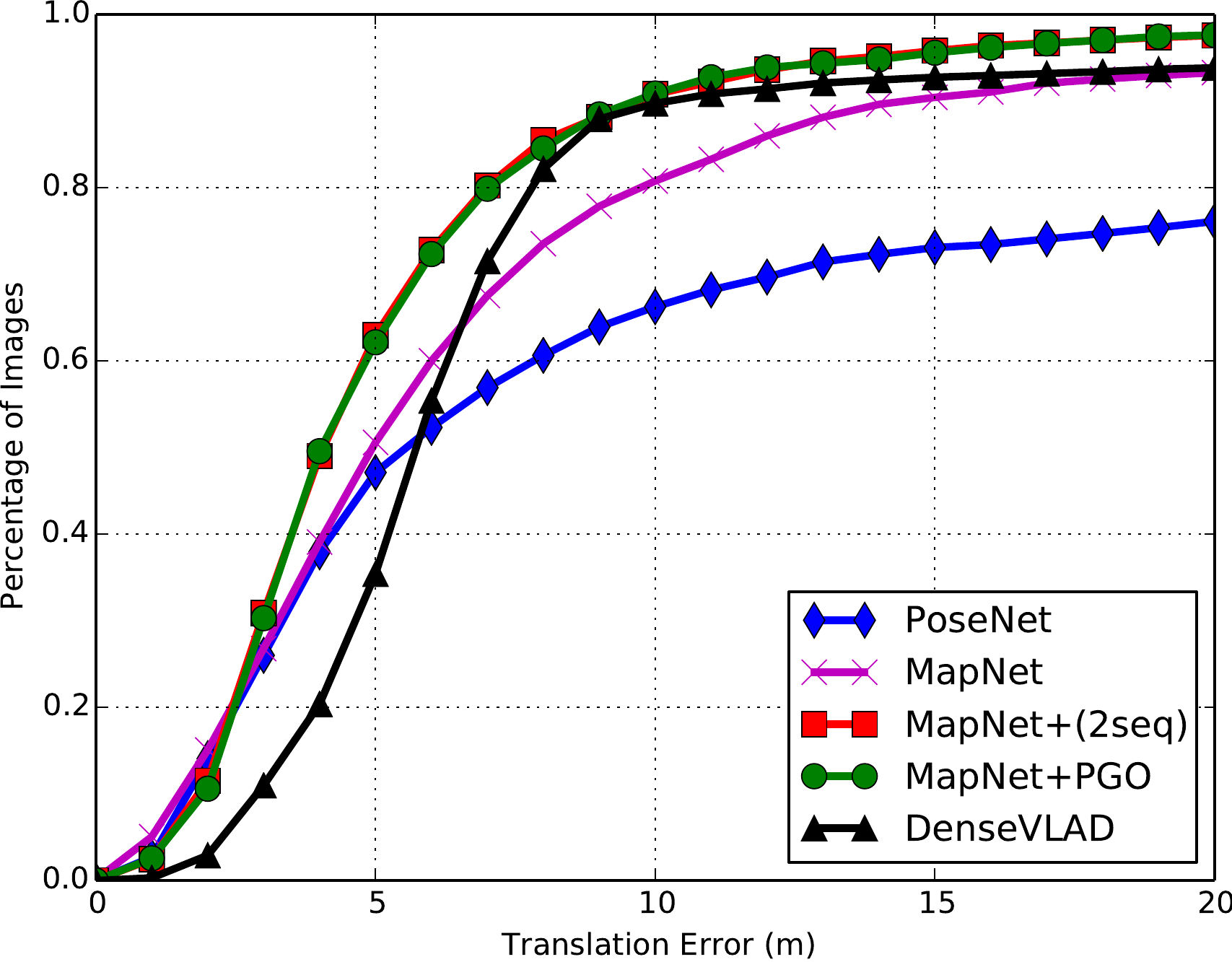}%
\hspace{2pt}%
\includegraphics[width=0.49\linewidth]{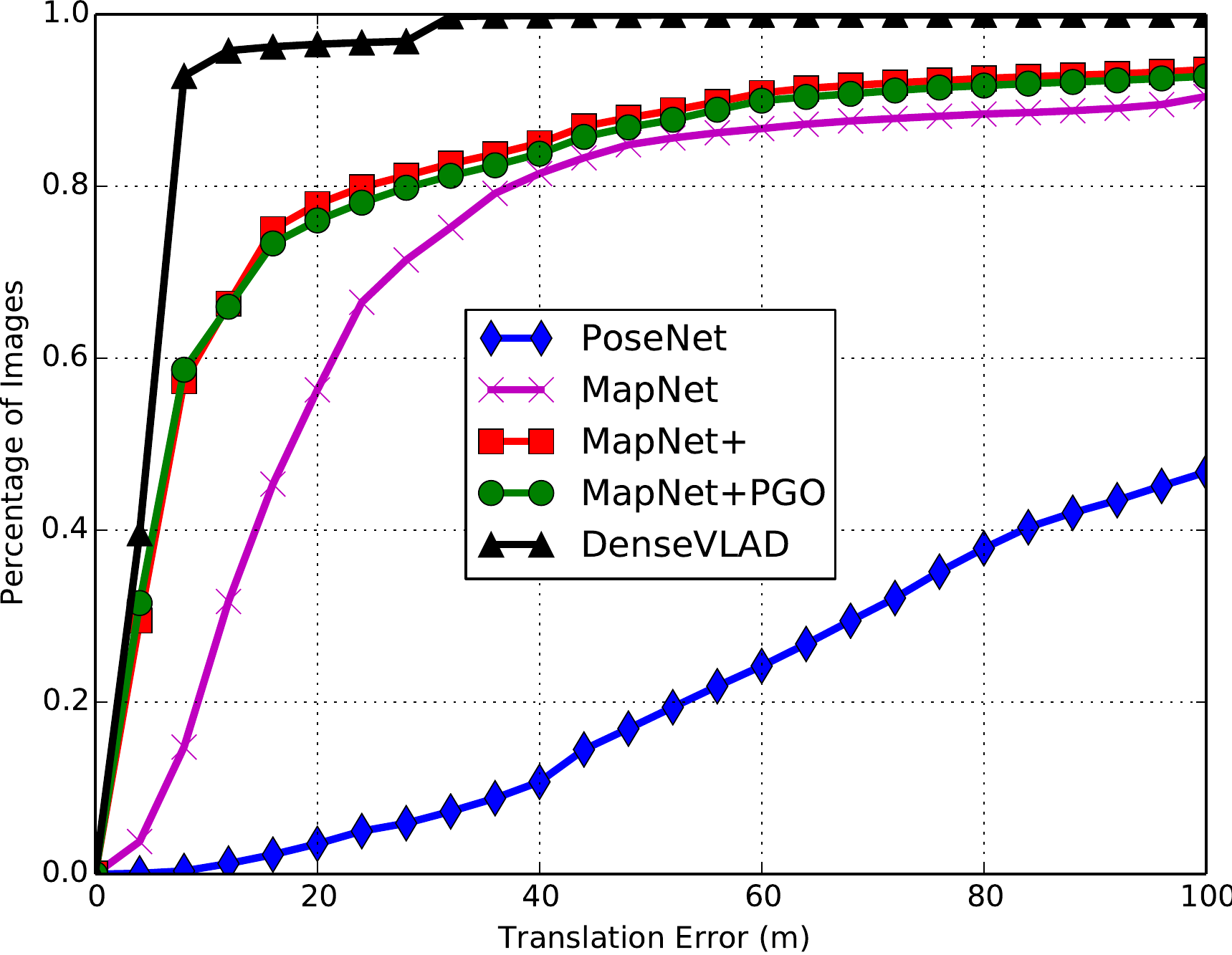}%
\end{center}
\vspace{-6pt}
\caption{Cumulative distribution of positional errors (in meters) for the (left) \textbf{RobotCar LOOP} and (right) \textbf{RobotCar FULL} datasets. On the larger dataset, DenseVLAD significantly outperforms pose regression techniques.}%
\label{fig:experiments:robotcar}
\end{figure}

\PAR{RobotCar dataset~\cite{Maddern2017IJRR}.} 
The training images of the LOOP and FULL scenes~\cite{Brahmbhatt2018CVPR} 
correspond to trajectories of 1.1km and 9.6km, respectively, driven by a car. 
The test images are obtained by driving the same trajectory. 
This dataset represents a scenario encountered during autonomous driving.

Fig.~\ref{fig:experiments:robotcar} shows the cumulative distributions of position errors for pose regression and image retrieval techniques.  
As expected, MapNet+ and MapNet+PGO outperform DenseVLAD on the smaller LOOP dataset. 
However, they perform significantly worse on the larger FULL scene\footnote{DenseVLAD is slightly more accurate on the FULL scene than on the LOOP dataset. We attribute this to the image quality, as the test set of the LOOP scene contains several overexposed images.}. 
This is despite MapNet+ using additional training sequences and MapNet+PGO using information from multiple images for its predictions. %
This scalability issue of pose regression is in line with similar observations in the literature~\cite{Sattler2018CVPR,Taira2018CVPR,Schoenberger2018CVPR}. 

\PAR{Using densely sampled data.} As in Sec.~\ref{sec:generalization}, our final experiment compares image retrieval and APR techniques on a synthetic scene, where a vast amount of training data is available. %
As shown in Tab.~\ref{tab:experiments:synthetic}, MapNet outperforms the image retrieval baselines when more training data is available. 
Still, it performs much closer to the retrieval baselines than to the structure-based method.

\section{Conclusion}
In this paper, we have derived a theoretic model for absolute pose regression (APR) algorithms. 
For the first time, this model allowed us to develop a better understanding of what APR method are and  are not capable of. 
Based on our theory, we have predicted that APR techniques are not guaranteed to generalize from the training data in practical scenarios. 
We have also shown that APR is more closely related to image retrieval approaches than to methods that accurately estimate camera poses via 3D geometry. 
These predictions %
have been verified through extensive experiments. 

The second main result of our paper is to show that pose regression techniques are currently competing with image retrieval approaches approximating the test pose rather than with methods that compute it accurately. 
More precisely, we have shown that no current pose regression approach consistently outperforms a handcrafted retrieval baseline. 
This paper has thus introduced an important sanity check for judging pose regression methods, showing that there is still a significant amount of research to be done before pose regression approaches become practically relevant. 

\small{
\PAR{Acknowledgements.}
This research was partially funded by the Humboldt Foundation through the Sofja Kovalevskaya Award.
}

\appendix
\section*{Appendix}
\label{sec:appendix}
The appendix consists of two parts: 
\textbf{i)} The accompanying video shows how the base translations estimated by MapNet~\cite{Brahmbhatt2018CVPR} are coupled to the image content  and illustrates the poses predicted for the test images in some of the scenes shown in the paper. 
The video is available at \url{https://github.com/tsattler/understanding_apr}.
Sec.~\ref{sec:appendix:video} gives a short overview over the video.
\textbf{ii)} Sec.~\ref{sec:appendix:add_experiment} presents an additional experiment on the DeepLoc dataset~\cite{Radwan2018RAL} that was left out of the paper due to space constraints.

\section{Supplementary Video}
\label{sec:appendix:video}
The video consists of two parts: 
The first part shows how the impact of each estimated base translation on the predicted pose depends on the image content. 
This is shown for the training images from the scene from Fig.~\ref{fig:lin_motion} (right).

The second part shows the positions estimated for the test images. 
We show the test image itself, the most similar training image (where similarity is measured based on the embeddings in the high-dimensional space), the base translations for the two images, and a 2D top-down view of the camera trajectories. 
In the 2D view, we show the ground truth training and testing positions, the pose of the current test image predicted by an absolute pose regression technique, the ground truth pose of the test image, and the pose of the most similar training images. 

For all experiments shown in the video, the absolute pose regression technique used was MapNet~\cite{Brahmbhatt2018CVPR}. 
Only test images that can be localized by Active Search~\cite{Sattler2017PAMI} are shown.

\begin{table*}
\footnotesize{
\setlength{\tabcolsep}{2.5pt}
\begin{center}
\begin{tabular}{c|c|c|c|c|c|c|c|c}
Pose & Bay. & SVS & VLocNet & DenseVLAD & DenseVLAD & VLocNet++$_\text{STL}$ & VLocNet++$_\text{MTL}$ & Active \\ 
Net~\cite{Kendall2015ICCV} & PoseNet~\cite{Kendall2016ICRA} & Pose~\cite{Naseer2017IROS} & \cite{Valada2018ICRA} & \cite{Torii15CVPR} & +Inter. & \cite{Radwan2018RAL} & \cite{Radwan2018RAL} & Search~\cite{Sattler2017PAMI}\\ \hline
2.42m, 3.66$^\circ$ & 2.24m, 4.31$^\circ$ & 1.61m, 3.52$^\circ$ & 0.68m, 3.43$^\circ$ & 0.57m, 3.15 $^\circ$ & 0.48m, 3.14$^\circ$ & 0.37m, 1.93$^\circ$ & 0.32m, 1.48$^\circ$ & \\ \hline
& & & & 0.51m, 2.57$^\circ$ & 0.44m, 2.52$^\circ$  & & & 0.01m, 0.04$^\circ$
\end{tabular}
\end{center}
}
\caption{Median position and orientation errors on the \textbf{DeepLoc} dataset~\cite{Radwan2018RAL}. \emph{DenseVLAD+Inter.} uses the top-15 retrieved images for interpolation. We show results for (top row) the original dataset and (bottom) our SfM version of the dataset.}%
\label{tab:experiments:deeploc}
\end{table*}

\section{Experiments on the DeepLoc Dataset~\cite{Radwan2018RAL}}
\label{sec:appendix:add_experiment}
\begin{figure}[t!]
\begin{center}
\includegraphics[width=\linewidth]{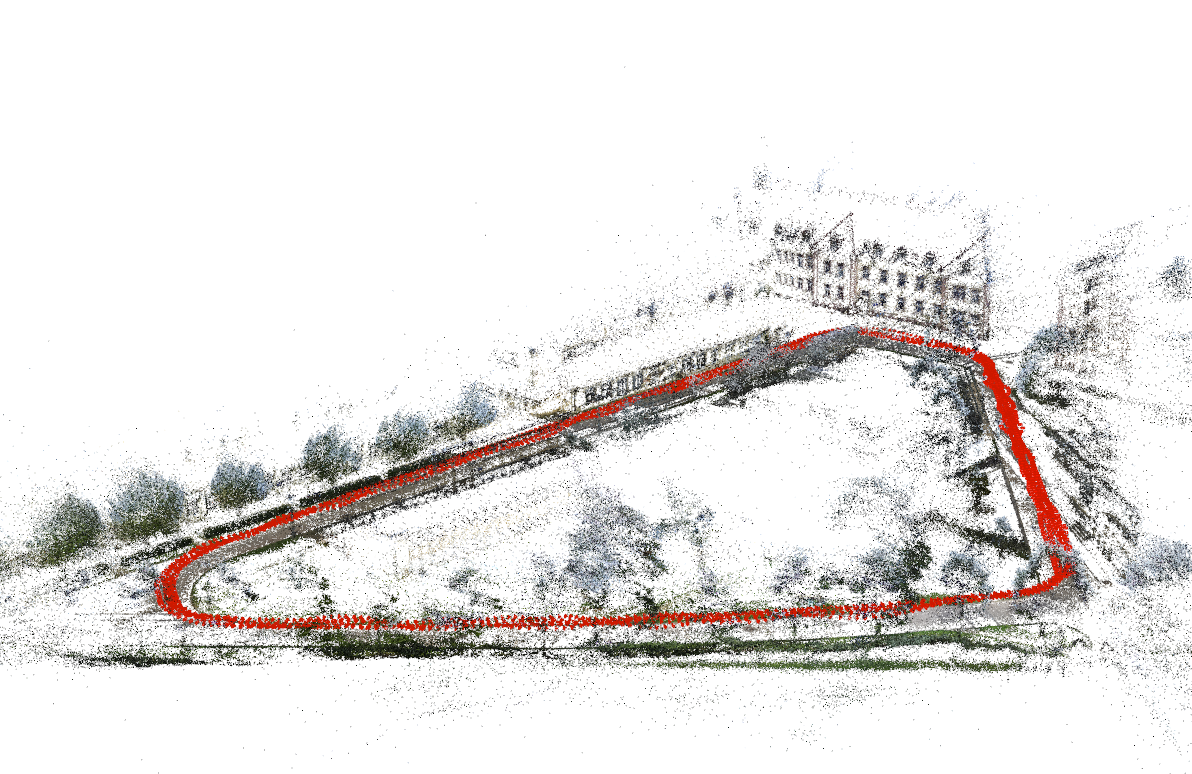}%
\end{center}%
\caption{Visualization of the SfM model of the DeepLoc dataset~\cite{Radwan2018RAL} that we constructed from the training images (red).}%
\label{fig:deeploc}
\end{figure}

The DeepLoc\footnote{\url{http://deeploc.cs.uni-freiburg.de/}} dataset~\cite{Radwan2018RAL} 
was captured from a robot driving a triangular-shaped trajectory multiple times (\cf Fig.~\ref{fig:deeploc}). 
In contrast to the RobotCar dataset~\cite{Maddern2017IJRR}, which was captured in an urban environment, %
the DeepLoc dataset shows a significant amount of vegetation. 

Tab.~\ref{tab:experiments:deeploc} (first row) compares the results obtained with  DenseVLAD~\cite{Torii15CVPR} without (\emph{DenseVLAD}) and with interpolation (\emph{DenseVLAD+Inter.}) with the results for various absolute pose regression techniques reported in~\cite{Radwan2018RAL}. %
Again, DenseVLAD significantly outperforms pose regression approaches based on a single image~\cite{Kendall2015ICCV,Kendall2016ICRA,Naseer2017IROS}. 
The table also compares DenseVLAD and DenseVLAD+Inter. against three sequence-based approaches, VLocNet~\cite{Valada2018ICRA}, VLocNet++$_\text{STL}$~\cite{Radwan2018RAL},  and VLocNet++$_\text{MTL}$~\cite{Radwan2018RAL}. 
All three directly fuse feature map responses from the previous time step $t-1$ into the CNN that predicts the pose at time $t$. 
VLocNet++$_\text{MTL}$ also integrates some form of higher-level scene understanding through semantic segmentation. %
All three methods operate on image sequences and thus use more information compared to DenseVLAD, which only uses a single image for localization. 
Still, DenseVLAD outperforms VLocNet~\cite{Valada2018ICRA}. %

The ground truth for the DeepLoc dataset was created using LIDAR-based SLAM. 
The dataset only provides the poses of the LIDAR sensor and not the cameras. 
This is not an issue for pose regression techniques as the camera and the LIDAR are related by a fixed (but unknown) transformation and it is irrelevant for the regressor which of the two local coordinate systems is used. 
However, not knowing the relative transformation from the LIDAR to the camera coordinate system  prevents us from easily creating a 3D model for structure-based methods. %
In order to be able to compare against Active Search~\cite{Sattler2017PAMI}, we thus created a second version of the dataset using SfM~\cite{Schoenberger2016CVPR}. 
To this end, we ran SfM on both the training and test images together. 
We then registered the SfM model against the LIDAR ground truth poses\footnote{There seems to be some drift in the vertical direction for the LIDAR poses while there seems to be little height variation in the scene. We thus use a variant of the original ground truth positions, where all heights are set to the same value, for computing the alignment between the SfM model and the positions.} to recover the scale of the model. 
This provided us with ground truth poses for the training and test images.
Finally, we used the ground truth poses of the training images and the feature matches between them to triangulate the 3D model used by Active Search\footnote{As was done for the datasets used in the paper.}. 
This ensures that the 3D model used for localization only contains information from the training images.

The second row of Tab.~\ref{tab:experiments:deeploc} shows the results obtained by Active Search on our version of the dataset. 
As can be seen, Active Search is significantly more accurate than all pose regression techniques, including VLocNet++$_\text{MTL}$, even though it only uses a single image for localization. 
For reference, we also include results obtained with DenseVLAD and DenseVLAD+Inter. on this new version of the dataset. 
As can be seen, the results obtained via DenseVLAD and DenseVLAD+Inter. do not change significantly between both versions of the datasets. 
This shows that the results obtained by Active Search and the pose regression algorithms on the two variants of the dataset are comparable. 

{\small
\bibliographystyle{ieee}
\bibliography{pose-cvpr19.bib}
}

\end{document}